\documentclass[conference]{IEEEtran}
\IEEEoverridecommandlockouts
\usepackage{cite}
\usepackage{amsmath,amssymb,amsfonts}
\usepackage{algorithmic}
\usepackage{graphicx}
\usepackage{textcomp}
\usepackage{xcolor}
\usepackage{url}

\def\BibTeX{{\rm B\kern-.05em{\sc i\kern-.025em b}\kern-.08em
    T\kern-.1667em\lower.7ex\hbox{E}\kern-.125emX}}
\begin{document}

\title{Network-Efficient World Model Token Streaming\\
}
\newcommand{\aff}[1]{\textsuperscript{#1}}
\author{
\IEEEauthorblockN{
Shatadal Mishra,
Ahmadreza Moradipari,
Nejib Ammar
}
\IEEEauthorblockA{InfoTech Labs, Toyota Motor North America R\&D, Mountain View, CA, USA}
\IEEEauthorblockA{\{shatadal.mishra, ahmadreza.moradipari, nejib.ammar\}@toyota.com}
}

\maketitle

\begin{abstract}
Generative driving world models rely on compact latent state representations that must be efficiently transmitted and synchronized across distributed compute and connected vehicles. We study network-efficient streaming of a \emph{discrete} world model state, where a stride-16 VQ-U-Net tokenizer (codebook size 8{,}192) maps each 288$\times$512 frame to an 18$\times$32 grid of token IDs (576 tokens/frame), equivalent to 936 bytes/frame under fixed-length coding. We consider a keyframe--delta protocol under strict per-message payload budgets and packet loss, and propose a fully online, label-free algorithm that prioritizes delta updates via cosine distance in codebook embedding space and triggers keyframes adaptively using a Hamming-drift threshold.
The adaptive algorithm consistently improves the rate distortion frontier over periodic keyframes at matched bitrates: at $\sim$0.024 \,Mb/s (200-byte budget) dynamic-only embedding distortion drops from 0.0712 to 0.0661 (7.2\%), and at $\sim$0.036 \,Mb/s (400-byte budget) from 0.0427 to 0.0407 (4.8\%). Under 10\% delta packet loss at 200 bytes, dynamic-only distortion is 0.0757 versus 0.0789 for a matched periodic baseline. To connect state fidelity to world model usefulness, we train a lightweight next-token predictor and evaluate perplexity conditioned on streamed receiver states: at $\sim$0.024\,Mb/s, dynamic-position perplexity improves from 206.0 to 193.1 (6.3\%), and at $\sim$0.036\,Mb/s from 158.9 to 155.6 (2.1\%). These results support discrete token-state streaming as a practical systems layer for bandwidth-aware synchronization and improved downstream token-dynamics utility under vehicular networking constraints.

\end{abstract}

\begin{IEEEkeywords}
world model, vehicular networking, cooperative driving, packet loss
\end{IEEEkeywords}

\section{Introduction}
Generative world models are emerging as a promising foundation for automated driving because they enable scalable simulation, long-horizon prediction, and closed-loop development within a unified learned framework \cite{b1,b2,b6}. By learning a predictive model of how the driving scene evolves, world models can support planning, counterfactual rollouts, and large-scale evaluation without requiring hand-engineered simulators \cite{b1,b2,b20}. However, a persistent practical barrier is representation: raw camera observations are high-dimensional, expensive to model directly, and awkward to store, transmit, or synchronize across distributed compute.

Modern driving world model pipelines therefore rely on compact latent states that preserve driving-relevant structure while reducing dimensionality \cite{b2,b6}. Discrete tokenization via vector quantization is especially attractive because it converts each frame into a spatial grid of discrete indices from a finite vocabulary \cite{b3}. Beyond the original VQ-VAE formulation, subsequent large-scale tokenized generation systems demonstrate that discrete codes can serve as a practical ``codec'' interface for powerful generative priors \cite{b12,b21}. This discrete interface has also been validated in widely used image generation pipelines that separate a learned tokenizer (or autoencoder) from a learned generative model operating in the compressed representation \cite{b4,b5,b14}. For video, tokenized representations have similarly enabled autoregressive modeling over spatiotemporal discrete latents \cite{b13}. In the driving domain, GAIA-1 adopts the same separation: a tokenizer compresses observations into a compact token state, a dynamics model operates in token space, and a renderer reconstructs pixels for visualization \cite{b6}. These trends suggest that discrete tokens are not merely a compression trick but a systems-level abstraction that makes high-capacity world models more tractable.

While prior work establishes the value of tokens for modeling and generation \cite{b3,b4,b5,b6,b12,b13}, an underexplored dimension is \emph{state streaming} under network constraints. Connected driving stacks increasingly span heterogeneous compute and communications resources: on-board inference, roadside/edge aggregation, and cloud backends for logging, evaluation, and large-scale learning \cite{b7,b8}. In such settings, state must be transmitted under strict constraints on message size, update rate, latency, and reliability \cite{b7,b8}. Related work in cooperative and collaborative perception underscores the fundamental trade-off between performance and communication bandwidth, and motivates careful budgeting of what to transmit \cite{b18,b17}. Even when a token grid is far smaller than pixels, naively sending the full state at high frequency remains inefficient and brittle: packet loss can de-synchronize the receiver state estimate, and multi-agent or multi-camera scaling can quickly saturate link budgets \cite{b7,b8}.

This paper studies \emph{network-efficient streaming and synchronization of discrete token states} for driving world models \cite{b3,b6}. Using a stride-16 VQ-U-Net tokenizer with a codebook of 8{,}192 entries, each $288\times512$ frame is mapped to an $18\times32$ grid of token IDs (576 tokens/frame). Under fixed-length coding (13 bits/token), this corresponds to 936 bytes/frame; at 10\,Hz, transmitting the full state would require roughly 0.075\,Mb/s per stream before headers and retransmissions. Our goal is to preserve state fidelity and downstream predictive usefulness while operating under per-message payload budgets (e.g., 100-800 bytes at 10\,Hz) and delta packet loss.

Our approach follows the classical keyframe-delta principle from video coding---periodic full refreshes complemented by lighter inter-frame updates \cite{b9,b15,b16,b11}---but operates directly in token space rather than pixels. The token stream provides a structured, discrete state that is naturally packetized and supports simple receiver-side state updates without a full pixel-domain codec pipeline. We propose a fully online, label-free algorithm with two components. First, for delta messages we prioritize which token positions to transmit by ranking changed positions using \emph{cosine distance in the codebook embedding space}; this distinguishes small, near-neighbor token substitutions from large, representation-changing substitutions. Second, we trigger keyframes \emph{adaptively} using a Hamming-drift threshold that measures divergence between the sender's current token state and the receiver's estimated state, enabling faster re-synchronization when drift spikes. This design requires no object detectors or semantic labels and is compatible with packet loss and strict message-size budgets.

We evaluate on 20\,s driving clips sampled at 10\,Hz under payload budgets and delta packet drops, reporting rate-distortion curves via \emph{dynamic-only embedding distortion}, robustness under packet loss, and algorithm behavior (keyframes-per-clip vs.\ drift threshold). Across 2{,}000 randomly sampled clips, adaptive keyframes improve the rate-distortion frontier over periodic keyframes at matched bitrates (e.g., 0.0712$\rightarrow$0.0661 at $\sim$0.024\,Mb/s and 0.0427$\rightarrow$0.0407 at $\sim$0.036\,Mb/s) and remain more robust under loss (10\% drops at 200 bytes: 0.0757 vs.\ 0.0789). To connect fidelity to world model usefulness, we train a lightweight next-token predictor \cite{b5,b6,b13} and observe lower dynamic-position perplexity at comparable bitrate (206.0$\rightarrow$193.1 at $\sim$0.024\,Mb/s; 158.9$\rightarrow$155.6 at $\sim$0.036\,Mb/s).

In summary, this work makes three contributions aimed at practical, connected deployments of driving world models:
\begin{enumerate}
    \item A packet-friendly discrete token state streaming design based on keyframes and budgeted delta updates, inspired by inter-frame coding principles \cite{b9,b15,b16,b11}.
    \item A fully online, label-free transmission algorithm that (i) prioritizes delta updates using cosine distance in token embedding space \cite{b3,b6,b12} and (ii) inserts keyframes adaptively based on Hamming drift under receiver-state divergence.
    \item An evaluation methodology for token-state synchronization under vehicular networking conditions \cite{b7,b8}, reporting dynamic-only state fidelity, robustness under packet loss, algorithm keyframe behavior, and downstream token-prediction utility \cite{b5,b6,b13}.
\end{enumerate}
Together, these results position discrete token-state streaming as a practical systems layer for bandwidth-aware world model synchronization, enabling efficient state sharing and improved downstream token-dynamics utility under realistic vehicular networking constraints \cite{b7,b8,b17,b18}.

\begin{figure*}[t]
  \centering
  \includegraphics[width=\textwidth]{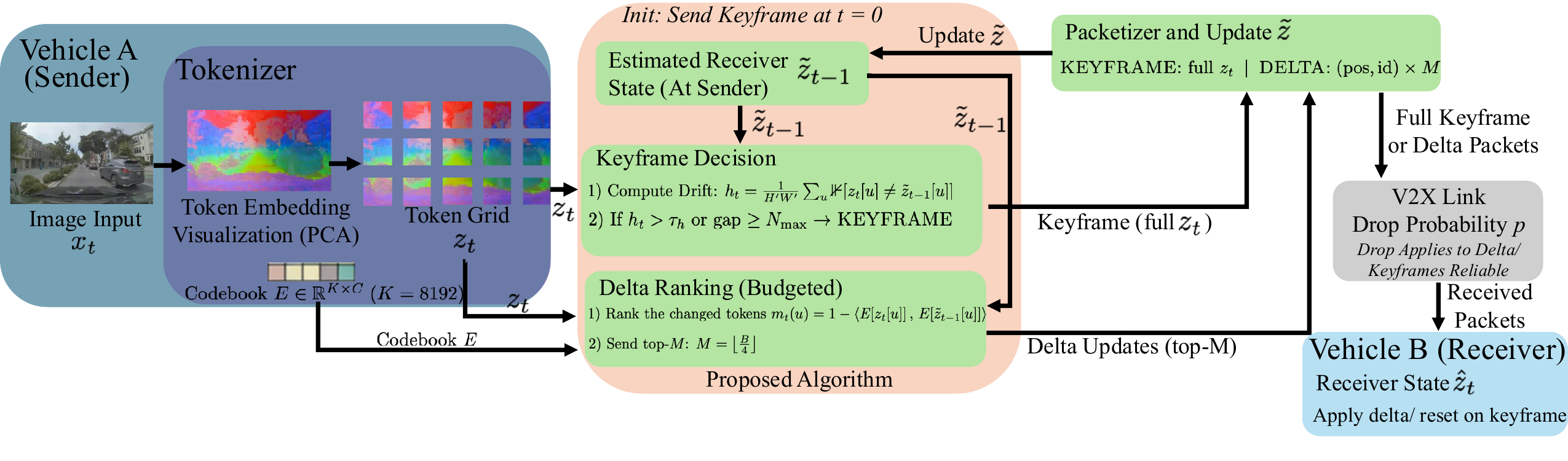}
  \caption{System architecture overview. Vehicle~A tokenizes each frame into a discrete state $z_t$ (18$\times$32, $K{=}8192$) and maintains a sender-side estimate $\tilde z_{t-1}$. Keyframes are triggered by Hamming drift ($\tau_h$, $N_{\max}$) and deltas transmit top-$M$ cosine-ranked updates using codebook embeddings $E$ under budget $B$; delta packets are dropped with probability $p$ (keyframes reliable). Vehicle~B reconstructs $\hat z_t$ by applying deltas or resetting on keyframes.}

  \label{fig:arch}
\end{figure*}

\section{System Model and Problem Statement}
We study synchronization of a \emph{tokenized world model state} over a bandwidth- and reliability-constrained vehicular link. A sender (Vehicle~A) observes a forward-facing camera stream and encodes each sampled frame into a spatial grid of discrete token IDs using a learned tokenizer. At time step $t$, the sender holds a token map
\begin{equation}
z_t \in \{0,\ldots,K-1\}^{H' \times W'},
\end{equation}
with vocabulary size $K=8192$ and grid resolution $H'\times W'=18\times 32$ (576 tokens per frame). In our experiments, tokens are extracted at a fixed update rate $R=10$\,Hz by uniform subsampling from 30\,fps driving clips of approximately 20\,s duration.

\subsection{Sender-Receiver State Synchronization}
Vehicle~A transmits state updates to a receiver (Vehicle~B) that maintains an internal reconstruction $\hat{z}_t$ intended to track the sender's true state $z_t$. Messages are generated at the token rate $R$ and are constrained by a per-message payload budget $B$ (bytes). Because $z_t$ is a discrete grid, the sender can either transmit a full refresh or a subset of token updates.

\paragraph{Keyframe-delta protocol.}
We adopt a keyframe-delta structure. A \emph{keyframe} carries a complete token grid and resets the receiver state, while a \emph{delta message} carries only a subset of token updates relative to the receiver estimate. A periodic baseline sends keyframes every $N$ steps (fixed keyframe interval), and sends deltas in between. Our proposed method also uses deltas but triggers keyframes \emph{adaptively} based on an online drift measure (defined below) with an additional maximum interval $N_{\max}$ to cap keyframe spacing.

\subsection{Packetization and Loss Model}
Each delta update consists of a token position identifier and a new token ID. For the $18\times 32$ grid, positions can be indexed in $[0,575]$, while token IDs require $\log_2 K=13$ bits. We use a conservative fixed accounting of 4 bytes per updated token (position + token ID) and a fixed header cost, $b_{hdr} = 20$ bytes, per message. Under budget $B$, the sender can transmit at most
\begin{equation}
M = \left\lfloor \frac{B - b_{hdr}}{b_{\text{upd}}}\right\rfloor
\end{equation}
token updates per delta message, where $b_{\text{upd}}=4$ bytes in our evaluation. A keyframe corresponds to transmitting the full grid, which is 936 bytes per frame under fixed-length coding (576 tokens $\times$ 13 bits).

We model packet loss by dropping delta messages independently with probability $p$. Unless stated otherwise, keyframes are assumed reliably delivered. When a delta is dropped, the receiver applies no updates for that step and retains its previous reconstruction until subsequent deltas and/or the next keyframe arrive.

\subsection{Receiver Reconstruction}
The receiver updates $\hat{z}_t$ sequentially. If a keyframe is received at time $t$, the receiver sets $\hat{z}_t = z_t$. Otherwise, for a delta message, the receiver applies only the transmitted position updates to its prior state:
\begin{equation}
\hat{z}_t[u] =
\begin{cases}
z_t[u], & u \in \mathcal{S}_t \ \text{and delta delivered},\\
\hat{z}_{t-1}[u], & \text{otherwise},
\end{cases}
\end{equation}
where $u$ indexes token positions and $\mathcal{S}_t$ is the set of positions selected for transmission at time $t$ (with $|\mathcal{S}_t| \le M$).

\subsection{Online Delta Selection via Embedding-Space Change}
Token IDs index entries in a learned codebook embedding table $E \in \mathbb{R}^{K\times C}$ (in our tokenizer $C=384$). We assume embeddings are $\ell_2$-normalized, so cosine similarity is a dot product. At time $t$, the sender compares the current token at position $u$ to the receiver estimate (or sender's estimate of the receiver) at the same position. Define an embedding-space change magnitude
\begin{align}
m_t(u)
&= 1 - \cos\!\big(E[z_t[u]],\,E[\tilde{z}_{t-1}[u]]\big) \nonumber\\
&= 1 - \left\langle E[z_t[u]],\,E[\tilde{z}_{t-1}[u]] \right\rangle .
\end{align}
where $\tilde{z}_{t-1}$ denotes the sender's reference state for selection (e.g., a sender-side estimate of the receiver's state). The sender forms the candidate set of changed positions $\mathcal{C}_t=\{u: z_t[u]\neq \tilde{z}_{t-1}[u]\}$ and transmits the top-$M$ positions ranked by $m_t(u)$. This yields a fully online, label-free prioritization rule: not all token changes are treated equally, and larger embedding-space shifts are prioritized under tight budgets.

\subsection{Adaptive Keyframe Trigger via Hamming Drift}
To prevent unbounded divergence under limited budgets and packet loss, we trigger keyframes adaptively using an online drift statistic. Let
\begin{equation}
h_t \;=\; \frac{1}{H'W'} \sum_{u} \mathbf{1}\!\left[z_t[u] \neq \tilde{z}_{t-1}[u]\right]
\end{equation}
denote the fraction of positions where the sender state differs from the reference (a normalized Hamming drift). Here, $\mathbf{1}[\cdot]$ denotes the indicator function, equal to 1 when the condition is true and 0 otherwise. A keyframe is transmitted when $h_t$ exceeds a threshold $\tau_h$, or when the time since the last keyframe exceeds a cap $N_{\max}$. This provides a simple resynchronization mechanism that reacts to scene changes and accumulated drift while bounding keyframe frequency.

\subsection{State Fidelity Metrics}
We evaluate reconstruction fidelity in token space by comparing $\hat{z}_t$ to $z_t$ offline. Because driving scenes exhibit substantial ego-motion and background dynamics, we report metrics that focus on \emph{changing} content. Let $\Delta_t(u)=\mathbf{1}[z_t[u]\neq z_{t-1}[u]]$ indicate whether a position changes between consecutive ground-truth frames. We define:
\begin{itemize}
    \item Token mismatch rate: $\mathbb{E}_{t,u}\,\mathbf{1}[\hat{z}_t[u]\neq z_t[u]]$.
    \item Dynamic-only embedding distortion (primary):
    \begin{equation}
    D_{\text{dyn}} \;=\; \mathbb{E}_{t,u:\,\Delta_t(u)=1}\left[1-\cos\!\big(E[z_t[u]],E[\hat{z}_t[u]]\big)\right].
    \end{equation}
\end{itemize}
Embedding distortion distinguishes near-miss token errors from far errors in the representation consumed by token-space predictors.

\subsection{World Model Utility under Streaming Constraints}
Beyond state fidelity, a world model oriented protocol should preserve downstream predictive utility at the receiver. We train a lightweight next-token predictor and evaluate next-step prediction quality when conditioning on the reconstructed receiver history $\hat{z}$ instead of the ground truth $z$. Concretely, for context length $L$ the predictor models $p(z_t[u] \mid z_{t-L:t-1}[u],u)$, and we evaluate
\begin{equation}
\mathcal{L}_{\text{WM}} \;=\; \mathbb{E}_{t,u}\left[-\log p\!\left(z_t[u] \mid \hat{z}_{t-L:t-1}[u],u\right)\right],
\end{equation}
reported as cross-entropy / perplexity. We additionally report this utility on dynamic positions to align with the settings where synchronization errors most affect prediction.

\subsection{Problem Statement}
Given a token stream $\{z_t\}$ at rate $R$, a per-message payload budget $B$, and a delta-message drop probability $p$, we seek a streaming policy that determines (i) which positions $\mathcal{S}_t$ to transmit in each delta message, and (ii) when to send keyframes, to achieve favorable trade-offs among:
\begin{enumerate}
    \item Communication cost: bitrate and message size under the payload budget;
    \item State fidelity: low dynamic-only embedding distortion and mismatch rate;
    \item World Model utility: low next-token prediction loss/perplexity when conditioning on the reconstructed receiver state.
\end{enumerate}
Our evaluation instantiates this system end-to-end: Vehicle~A generates discrete token states, packetizes them into keyframes and deltas under a byte budget, Vehicle~B reconstructs $\hat{z}_t$ under packet loss, and a token predictor measures how synchronization quality impacts downstream predictive utility.

\section{Proposed Method}
We propose a feedback-free, packet-friendly protocol for discrete token-state synchronization under payload budgets and delta packet loss. The protocol combines keyframe--delta packetization with an online policy for budgeted delta selection and adaptive keyframe insertion.

\subsection{Keyframe-delta Token-State Streaming}
Let the sender’s tokenized world model state at time $t$ be
$z_t \in \{0,\ldots,K-1\}^{H' \times W'}$ with $H' \times W' = 18 \times 32$ and
$K=8192$. We flatten the grid into $N_{\text{pos}} = H'W' = 576$ positions indexed by
$u \in \{0,\ldots,N_{\text{pos}}-1\}$, and denote the token at position $u$ by $z_t[u]$.

The sender transmits a sequence of messages at the token rate (10\,Hz in our experiments).
A \emph{keyframe} message carries the complete token grid $z_t$ and resets the receiver state.
A \emph{delta} message carries only a subset of token updates for positions whose current token differs
from a sender-side reference state. This mirrors keyframe/inter-frame coding principles, but operates
directly in discrete token space. We always transmit an initial keyframe at $t=0$ to initialize the receiver state, i.e., $\hat{z}_0 \leftarrow z_0$ and $\tilde{z}_0 \leftarrow z_0$; subsequent timesteps use delta updates and keyframe decisions as described below.

\paragraph{Sender-side reference state.}
In the absence of acknowledgments, the sender maintains a reference $\tilde{z}_{t-1}$
that approximates the receiver’s current state. In our implementation, $\tilde{z}_{t-1}$
is updated optimistically by applying the updates that the sender transmits (i.e., assume delivered).
When a keyframe is sent, the reference is refreshed:
$\tilde{z}_t \leftarrow z_t$.

\subsection{Update Encoding and Byte Budget}
A delta message contains a list of updates $(u, z_t[u])$.
For $N_{\text{pos}}=576$ positions, the position index requires 10 bits, and for $K=8192$,
the token ID requires 13 bits. While the theoretical minimum is 23 bits/update, practical packet formats
typically use fixed-width fields. We therefore adopt a conservative accounting of $b_{\text{upd}}=4$ bytes per update
(position + token ID), plus a fixed header cost $b_{\text{hdr}}$ per message.
Given a per-message payload budget $B$ bytes, the maximum number of token updates that can be transmitted in a delta follows the byte budget model defined in (2).

\subsection{Budgeted Delta Selection via Embedding-Space Change}
Sending all changed positions is generally infeasible under tight budgets. Our core design choice is to prioritize
which changed tokens to transmit using a \emph{continuous} measure of how large the change is in the tokenizer’s
codebook embedding space.

Let $E \in \mathbb{R}^{K \times C}$ be the learned codebook embedding table (with $C=384$ in our tokenizer).
We use $\ell_2$-normalized embeddings, so cosine similarity reduces to a dot product.
At time $t$, define the candidate set of positions that differ from the reference:
\begin{equation}
\mathcal{C}_t \;=\; \{\, u \mid z_t[u] \neq \tilde{z}_{t-1}[u] \,\}.
\end{equation}
For each candidate position $u \in \mathcal{C}_t$, we compute an embedding-space change magnitude
\begin{align}
m_t(u)
&= 1 - \cos\!\big(E[z_t[u]],\,E[\tilde{z}_{t-1}[u]]\big) \nonumber\\
&= 1 - \left\langle E[z_t[u]],\,E[\tilde{z}_{t-1}[u]] \right\rangle .
\end{align}
The sender ranks all $u \in \mathcal{C}_t$ by $m_t(u)$ and transmits the top-$M$ updates
\begin{equation}
\mathcal{S}_t \;=\; \operatorname{TopM}\left(\{(u, m_t(u))\}_{u\in\mathcal{C}_t}\right), \qquad |\mathcal{S}_t| \le M.
\end{equation}
Intuitively, this rule treats token changes as unequal: near-neighbor substitutions in embedding space (small $m_t$)
are deprioritized relative to large representation shifts (large $m_t$), under the same byte budget.

\textit{Computational overhead}:
The proposed delta-ranking step is lightweight because it operates only on the $18 \times 32$ token grid. At each timestep, it requires at most $576 \times 384$ dot-product operations over $\ell_2$-normalized codebook embeddings, followed by a top-$M$ selection. At 10 Hz, this cost is small compared with tokenizer inference and does not require object detection, semantic segmentation, or a learned scheduler. End-to-end deployment should still account for tokenizer latency, packetization, and network delay.

\subsection{Adaptive Keyframe Trigger via Hamming Drift}
To prevent unbounded divergence when only sparse deltas are transmitted (and when deltas are dropped), we trigger keyframes
adaptively using an online drift statistic. Define a normalized Hamming drift between the sender state and the reference:
\begin{equation}
h_t \;=\; \frac{1}{N_{\text{pos}}}\sum_{u=0}^{N_{\text{pos}}-1} \mathbf{1}\!\left[z_t[u] \neq \tilde{z}_{t-1}[u]\right].
\end{equation}
A keyframe is transmitted at time $t$ if either (i) drift exceeds a threshold $\tau_h$, or (ii) a maximum keyframe gap is reached:
\begin{equation}
\text{send keyframe at } t \ \ \text{if} \ \ (h_t > \tau_h)\ \ \text{or}\ \ (t - t_{\text{kf}} \ge N_{\max}),
\end{equation}
where $t_{\text{kf}}$ denotes the timestep at which the most recent full keyframe was transmitted and used to refresh the sender-side reference state. This drift-triggered insertion reacts to abrupt scene changes
and accumulated desynchronization, while $N_{\max}$ bounds keyframe spacing.

\subsection{Receiver Reconstruction and Loss Handling}
The receiver maintains a reconstructed state $\hat{z}_t$. Upon receiving a keyframe, it sets $\hat{z}_t = z_t$.
Upon receiving a delta message, it applies only the transmitted updates:
\begin{equation}
\hat{z}_t[u] =
\begin{cases}
z_t[u], & u \in \mathcal{S}_t \ \text{and delta delivered},\\
\hat{z}_{t-1}[u], & \text{otherwise}.
\end{cases}
\end{equation}
We model packet loss as independent drops of delta messages with probability $p$; unless noted, keyframes are assumed reliably delivered.
Under loss, the receiver may diverge from the sender reference $\tilde{z}_t$, but periodic or adaptive keyframes provide resynchronization. Figure~\ref{fig:arch} summarizes the end-to-end pipeline: sender tokenizes each frame into a discrete token grid $z_t$, uses a sender-side estimate $\tilde z_{t-1}$ to trigger keyframes (drift threshold $\tau_h$, cap $N_{\max}$) and select top-$M$ cosine-ranked delta updates under budget $B$, and transmits packets over a lossy V2X link (delta drop probability $p$) to receiver which reconstructs $\hat z_t$. 
We evaluate synchronization quality using dynamic-only embedding distortion between $z_t$ and $\hat z_t$, and downstream usefulness via next-token predictor perplexity conditioned on the reconstructed history $\hat z_{t-L:t-1}$.

\subsection{Baselines and Controllable Parameters}
Our evaluation compares the proposed adaptive algorithm against periodic keyframes (fixed interval $N$) with the same cosine-based delta ranking.
The method exposes a small set of interpretable parameters that trade bitrate, fidelity, and robustness:
\begin{itemize}
    \item payload budget $B$ and update cost $b_{\text{upd}}$ (bytes/update), header cost $b_{\text{hdr}}$;
    \item periodic keyframe interval $N$ (baseline);
    \item adaptive drift threshold $\tau_h$ and maximum gap $N_{\max}$ (proposed);
    \item delta message drop probability $p$.
\end{itemize}

\section{Experimental Setup}
We evaluate discrete token-state streaming on the NVIDIA driving dataset (front-wide camera), using $\sim$20\,s clips recorded at 30\,fps and uniformly subsampled to 10\,Hz ($\approx$200 timesteps/clip). Each sampled frame is encoded by a pretrained stride-16 VQ-U-Net tokenizer with codebook size $K{=}8192$, producing an $18\times32$ grid of token IDs (576 tokens/timestep). Under fixed-length coding (13 bits/token), transmitting the full grid corresponds to 936 bytes per timestep, which serves as an upper-bound bandwidth reference. Unless otherwise stated, curves are averaged over 2{,}000 randomly sampled clips, and stability is assessed via multiple random 500-clip subsets.

\subsection{Streaming Protocol and Configurations}
We simulate a sender-receiver pipeline in which the sender emits one message per timestep under a strict payload budget. Messages follow a keyframe-delta structure: keyframes transmit the complete token grid and reset the receiver state, while delta messages transmit only a subset of token updates.

\paragraph{Periodic baselines}
We evaluate periodic keyframe schedules that send a keyframe every $N$ timesteps and deltas otherwise, with
\[
N \in \{9,10,12,17,21\},
\]
chosen to span the bitrate range induced by our payload budgets at 10\,Hz and to provide rate-matched baselines for the adaptive algorithm.


\paragraph{Proposed adaptive algorithm}
For adaptive keyframes, we trigger a keyframe when the normalized Hamming drift $h_t$ (fraction of mismatched token positions) exceeds a threshold $\tau_h$, or when a maximum gap is reached ($N_{\max}=30$). We set $\tau_h$ from the empirical distribution of per-timestep token change rates measured on the dataset at 10\,Hz and sweep
\[
\tau_h \in \{0.814,\,0.757,\,0.727\},
\]
corresponding to the $99.9^{th}$, $99.5^{th}$, and $99^{th}$ percentiles. This is done to obtain strict-to-less-strict operating points that trade keyframe frequency against delta load. We choose high percentiles to make keyframes rare (avoiding keyframe spam) while still reacting to unusually large scene-change/drift events.

\paragraph{Payload budgets and update cost}
Delta messages are constrained by a per-message payload budget
\[
B \in \{100,\,200,\,400,\,800\}\ \text{bytes},
\]
and we use a conservative accounting of 4 bytes per token update (position + token ID) plus a fixed 20-byte header per message. Under budget $B$, the sender transmits at most $M=\lfloor (B-20)/4 \rfloor$ token updates per delta message. Keyframes are modeled as full-grid transmissions (936 bytes per timestep under fixed-length coding), plus the same fixed header cost.

\subsection{Packet Loss Model}
To evaluate robustness, we independently drop delta messages with probability
\[
p \in \{0.01,\,0.05,\,0.1\},
\]
while keyframes are assumed to be reliably delivered unless otherwise noted. When a delta message is dropped, the receiver does not apply updates for that timestep and continues in its previous state until a later delta is received successfully or a keyframe refresh occurs. This loss model isolates the effect of unreliable incremental updates and the recovery benefit of keyframes.

\subsection{Metrics: Rate, Fidelity, and Utility}
\paragraph{Communication cost}
We report average bitrate (Mb/s) computed from transmitted bytes and sequence duration. All methods transmit one message per timestep (10\,Hz), so the bitrate directly reflects the message size under the budget and keyframe schedule.

\paragraph{State fidelity (primary: dynamic-only)}
We evaluate fidelity in token space by comparing the receiver reconstruction $\hat{z}_t$ to the ground-truth sender state $z_t$. Because most positions can remain static for long periods, we emphasize dynamic content by reporting \emph{dynamic-only embedding distortion}:
\[
D_{\text{dyn}} \;=\; \mathbb{E}_{t,u:\,z_t[u]\neq z_{t-1}[u]}\left[1-\cos\!\big(E[z_t[u]],E[\hat{z}_t[u]]\big)\right],
\]
i.e., cosine distance between ground-truth and reconstructed codebook embeddings restricted to positions that change between consecutive ground-truth frames. We also track auxiliary metrics (e.g., mismatch fraction) for debugging and interpretability.
\begin{figure}[t]
  \centering
  \includegraphics[width=\columnwidth]{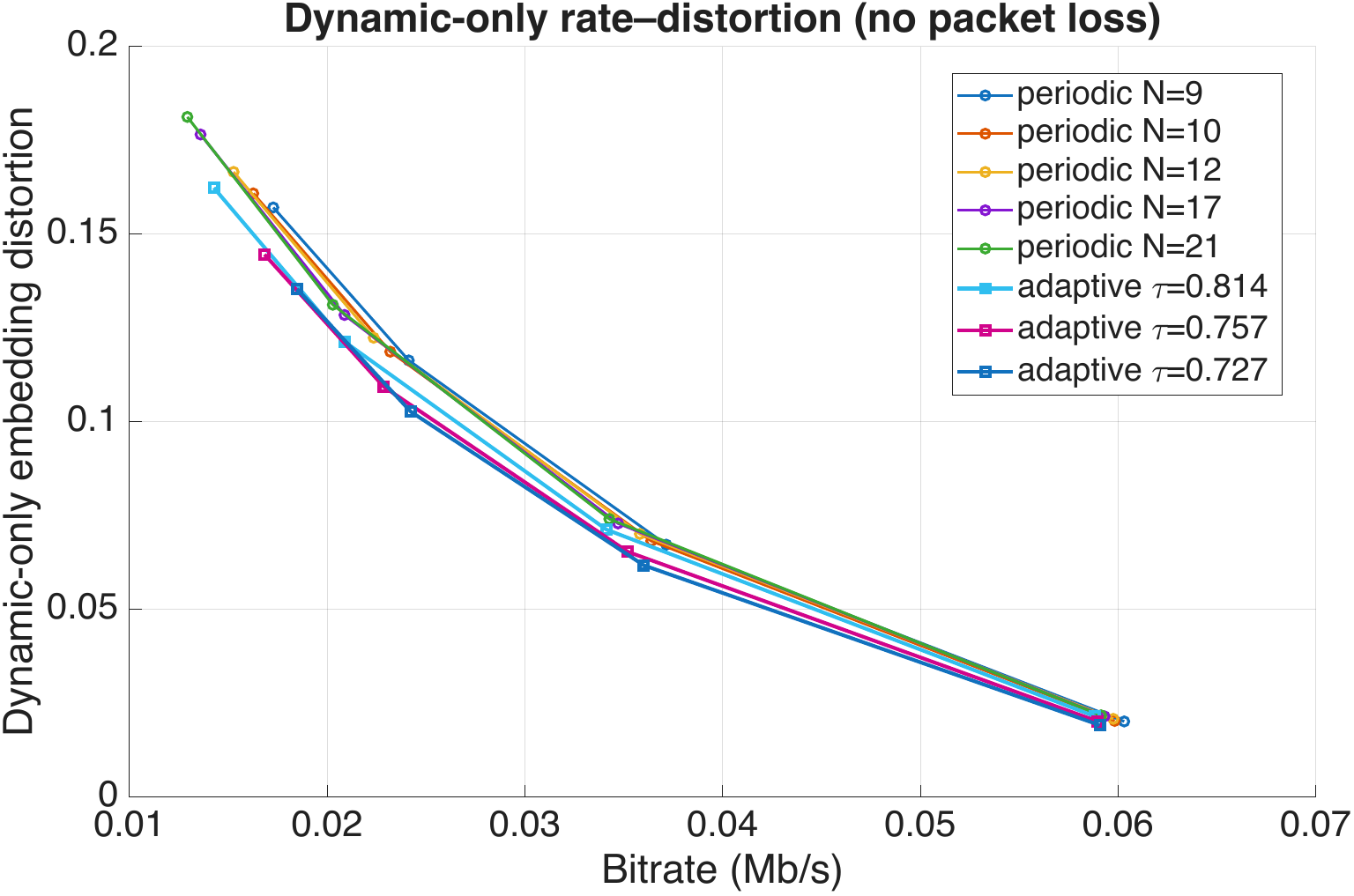}
  \caption{Dynamic-only rate-distortion (no packet loss). Bitrate vs dynamic-only embedding distortion (cosine distance between ground-truth and reconstructed codebook embeddings on positions that change) for 10\,Hz token-state streaming. Curves compare periodic keyframes ($N\in{9,10,12,17,21}$) and adaptive keyframes ($\tau_h\in{0.814,0.757,0.727}$); points are payload budgets $B\in{100,200,400,800}$ bytes (4 bytes/update). Lower is better.}
  \label{fig:matlab_dynamic_rd_p0}
\end{figure}

\paragraph{World model utility}
To connect synchronization quality with downstream predictive usefulness, we train a lightweight next-token predictor in token space and evaluate cross-entropy/perplexity when conditioning on the reconstructed history of the receiver. The predictor uses a context length $L=4$ and predicts $z_t[u]$ from $\hat{z}_{t-L:t-1}[u]$ and position $u$. We report perplexity over (i) all positions and (ii) dynamic positions, where dynamic-position perplexity is most sensitive to stale or missing updates. For efficiency, utility evaluation is performed on sampled timesteps and positions per clip (fixed across methods), while targets always remain the ground-truth next tokens.

\subsection{Stability and Significance Checks}
To ensure improvements are not an artifact of clip sampling, we perform repeated evaluations on multiple random clip subsets (10 seeds, 500 clips/seed) at rate-matched operating points. We report win rates and mean$\pm$std differences between the adaptive algorithm and matched periodic baselines for both fidelity (dynamic-only distortion) and utility (dynamic-position perplexity).
\begin{figure}[h]
  \centering
  \includegraphics[width=\columnwidth]{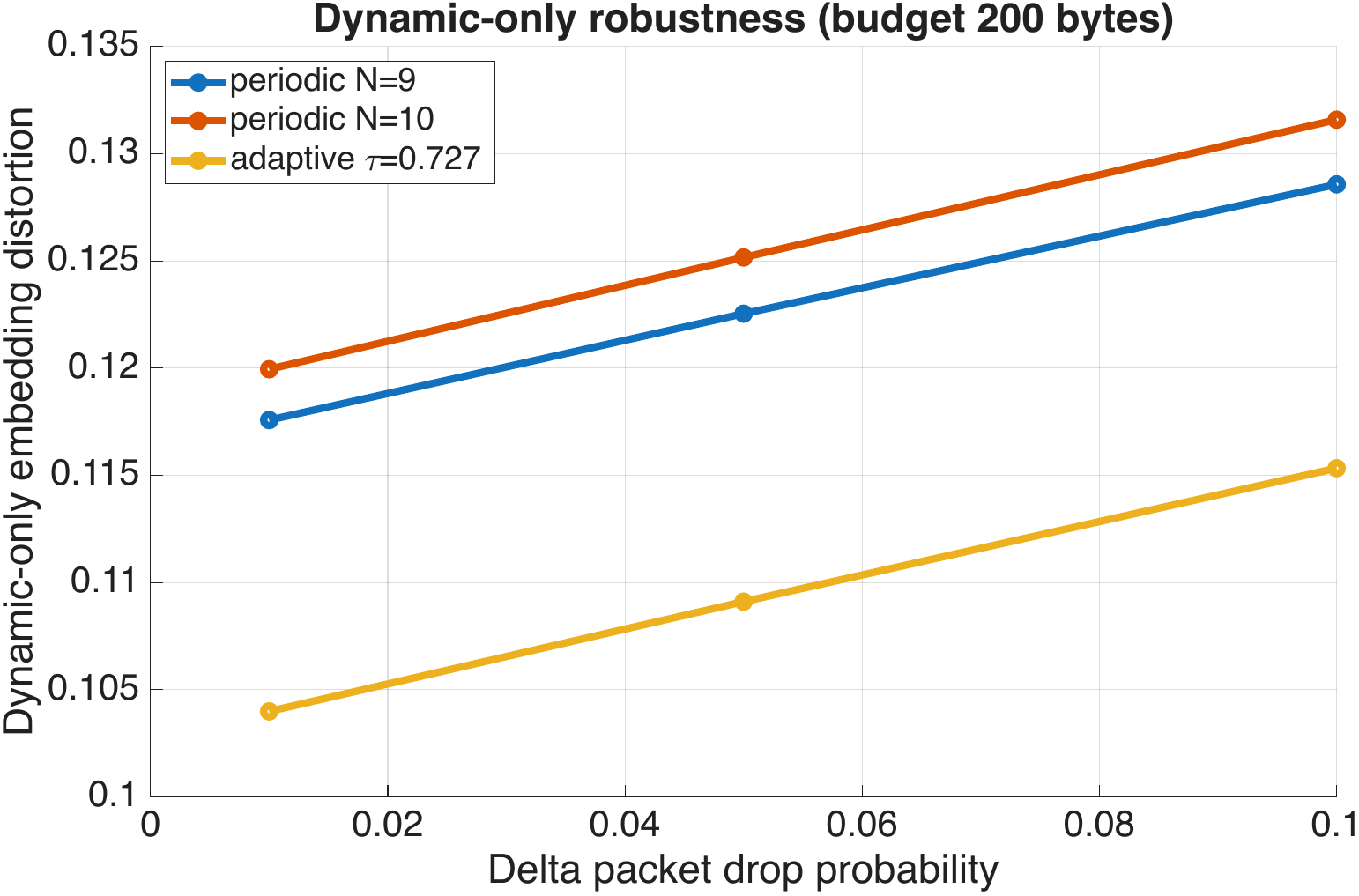}
  \caption{Dynamic-only robustness at 200-byte budget. Dynamic-only embedding distortion vs delta packet drop probability $p$ (10\,Hz). Periodic keyframes ($N=9,10$) are compared with the adaptive algorithm ($\tau_h=0.727$, $N_{\max}=30$), using 4 bytes/update and reliable keyframes. Lower is better.}
  \label{fig:matlab_fig2_dynamic_loss_budget200}
\end{figure}

\section{Simulation Results}
We evaluate network-efficient streaming of tokenized world-model states under payload budgets and delta packet loss. Unless otherwise stated, $\sim$20\,s clips are sampled at 10\,Hz, producing an $18\times 32$ (576 tokens) token grid per timestep. We report bitrate for communication cost, dynamic-only embedding distortion for state fidelity, robustness under independent delta drops, and next-token predictor perplexity for downstream world-model utility.

\subsection{Rate--Distortion Trade-offs without Packet Loss}
We first study the fundamental trade-off between bandwidth and state fidelity when there is no packet loss. Figure \ref{fig:matlab_dynamic_rd_p0} plots dynamic-only embedding distortion as a function of bitrate for (i) periodic keyframes with fixed intervals $N\in\{9,10,12,17,21\}$ and (ii) the proposed adaptive keyframe algorithm with drift thresholds $\tau_h \in \{0.814, 0.757, 0.727\}$. Across all settings, increasing the payload budget (100--800 bytes) increases bitrate and reduces distortion, reflecting that more delta updates per timestep lead to a closer receiver reconstruction.
\begin{figure}[b]
  \centering
  \includegraphics[width=\columnwidth]{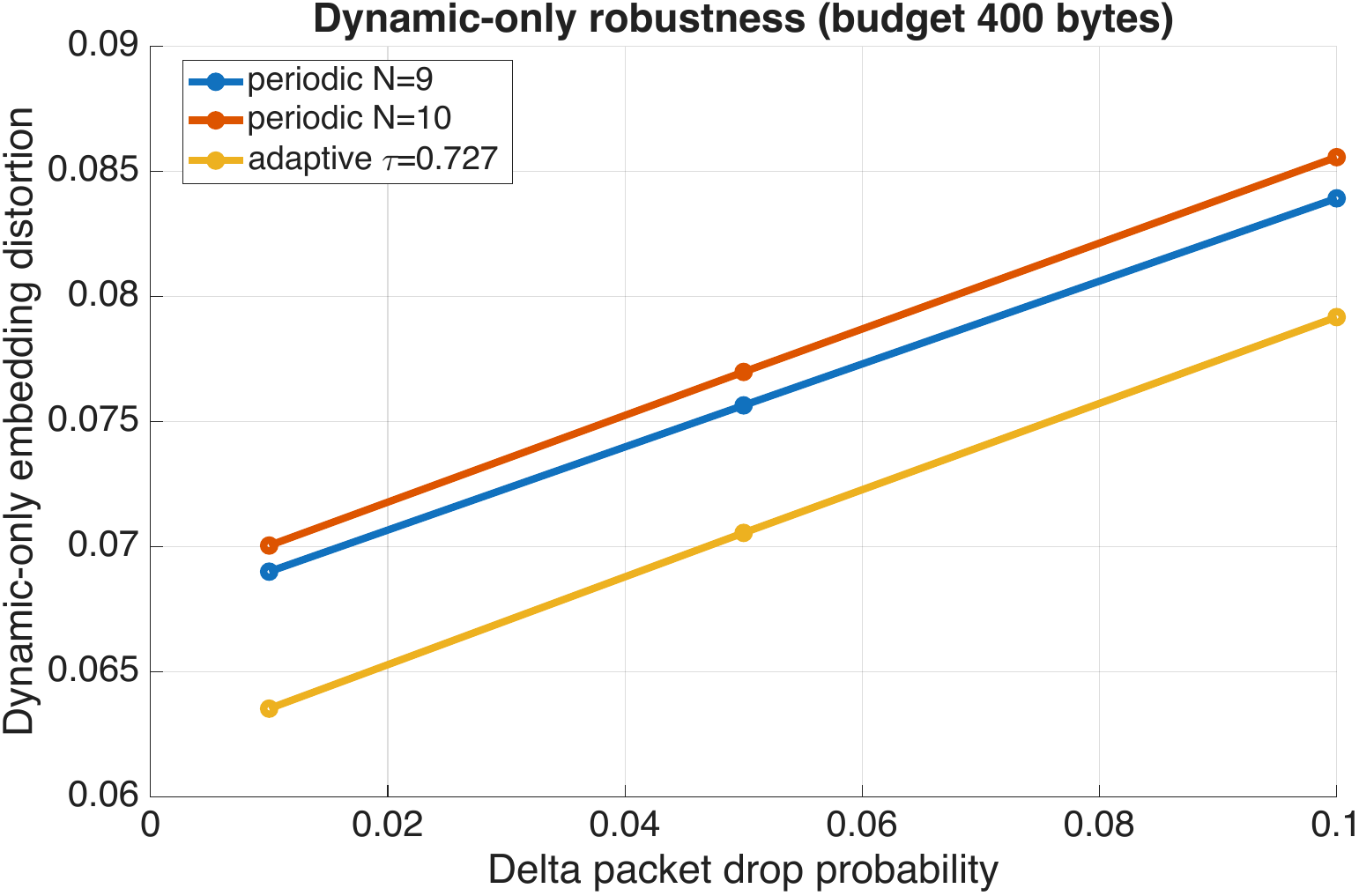}
  \caption{Dynamic-only robustness at 400-byte budget. Dynamic-only embedding distortion vs delta packet drop probability $p$ (10\,Hz). Periodic keyframes ($N=9,10$) are compared with the adaptive algorithm ($\tau_h=0.727$, $N_{\max}=30$), using 4 bytes/update and reliable keyframes. Lower is better.}
  \label{fig:matlab_fig3_dynamic_loss_budget400}
\end{figure}
A central result is that adaptive keyframes improve the rate--distortion frontier over periodic keyframes at matched bitrates in the operating regime relevant to tight budgets. For example, around the 200-byte operating point ($\sim$0.024\,Mb/s), adaptive keyframes achieve lower dynamic-only distortion (0.0661) than a matched periodic baseline (0.0712), corresponding to a 7.2\% relative reduction. At the 400-byte operating point ($\sim$0.036\,Mb/s), adaptive keyframes reduce distortion from 0.0427 to 0.0407 (4.8\% relative). At higher bitrates ($\sim$0.06\,Mb/s; 800 bytes), the curves converge, indicating diminishing returns as the receiver approaches full-state synchronization.

\subsection{Robustness to Delta Packet Loss}
We next evaluate robustness to delta-message drops. Figure \ref{fig:matlab_fig2_dynamic_loss_budget200} and Figure \ref{fig:matlab_fig3_dynamic_loss_budget400} report dynamic-only embedding distortion versus delta drop probability $p\in\{0.01,0.05,0.1\}$ for representative payload budgets (200 and 400 bytes, respectively). As expected, distortion increases monotonically with packet loss for all methods, since missed deltas cause the receiver state to become stale until a subsequent keyframe or successful delta update arrives. 

Importantly, adaptive keyframes remain consistently more robust at matched rates, maintaining lower distortion than periodic baselines across loss levels. For example, at 200 bytes and 10\% drop probability, adaptive keyframes achieve dynamic-only distortion 0.0757 compared to 0.0789 for the matched periodic baseline. At 400 bytes, the absolute gap is smaller but remains directionally consistent. These results indicate that drift-triggered keyframe insertion provides a simple and effective recovery mechanism that limits accumulated desynchronization under unreliable delta delivery.

\subsection{Algorithm Behavior: Keyframe Frequency vs Threshold}
\begin{figure}[t]
  \centering
  \includegraphics[width=\columnwidth]{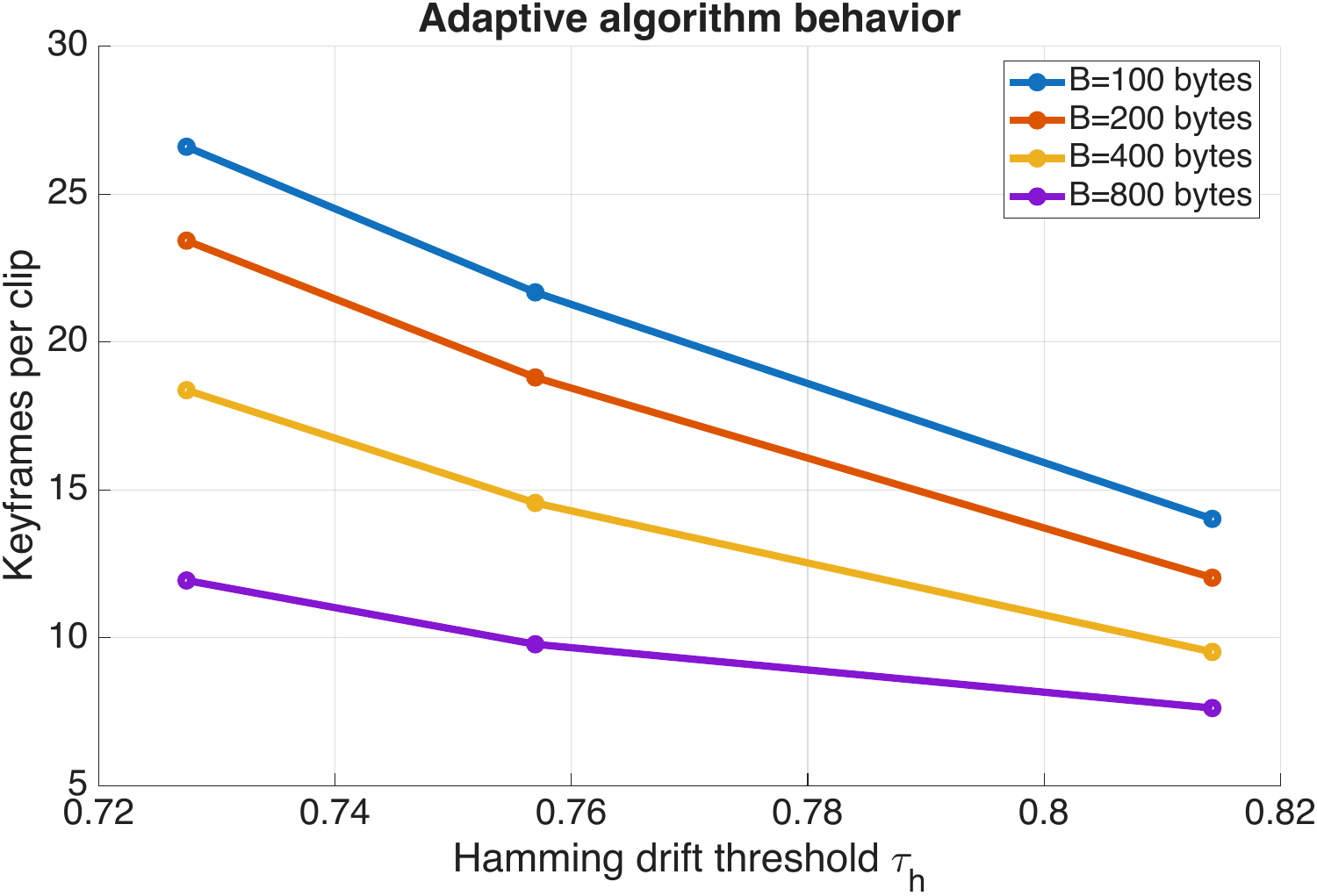}
  \caption{Adaptive keyframe behavior (no packet loss). Average keyframes per 20\,s clip (10\,Hz) versus Hamming drift threshold $\tau_h$ for delta payload budgets $B\in\{100,200,400,800\}$ bytes (4 bytes/update); larger $B$ and higher $\tau_h$ yield fewer keyframes.}
  \label{fig:matlab_fig4_keyframes_vs_tau}
\end{figure}

To interpret \emph{why} adaptive keyframes help, Figure \ref{fig:matlab_fig4_keyframes_vs_tau} plots keyframes-per-clip as a function of the drift threshold $\tau_h$, with separate curves for different payload budgets. Two trends are consistent across budgets. First, increasing $\tau_h$ reduces keyframe frequency, as the algorithm tolerates larger divergence before resetting. Second, increasing the payload budget reduces keyframe frequency at fixed $\tau_h$, since more delta updates are transmitted per timestep and the sender--receiver divergence grows more slowly. This figure provides an intuitive mechanism view: the adaptive algorithm spends keyframes selectively when drift spikes, and uses larger delta budgets to avoid unnecessary refreshes.

\subsection{World Model Utility under Streaming Constraints}
\begin{figure}[t]
  \centering
  \includegraphics[width=\columnwidth]{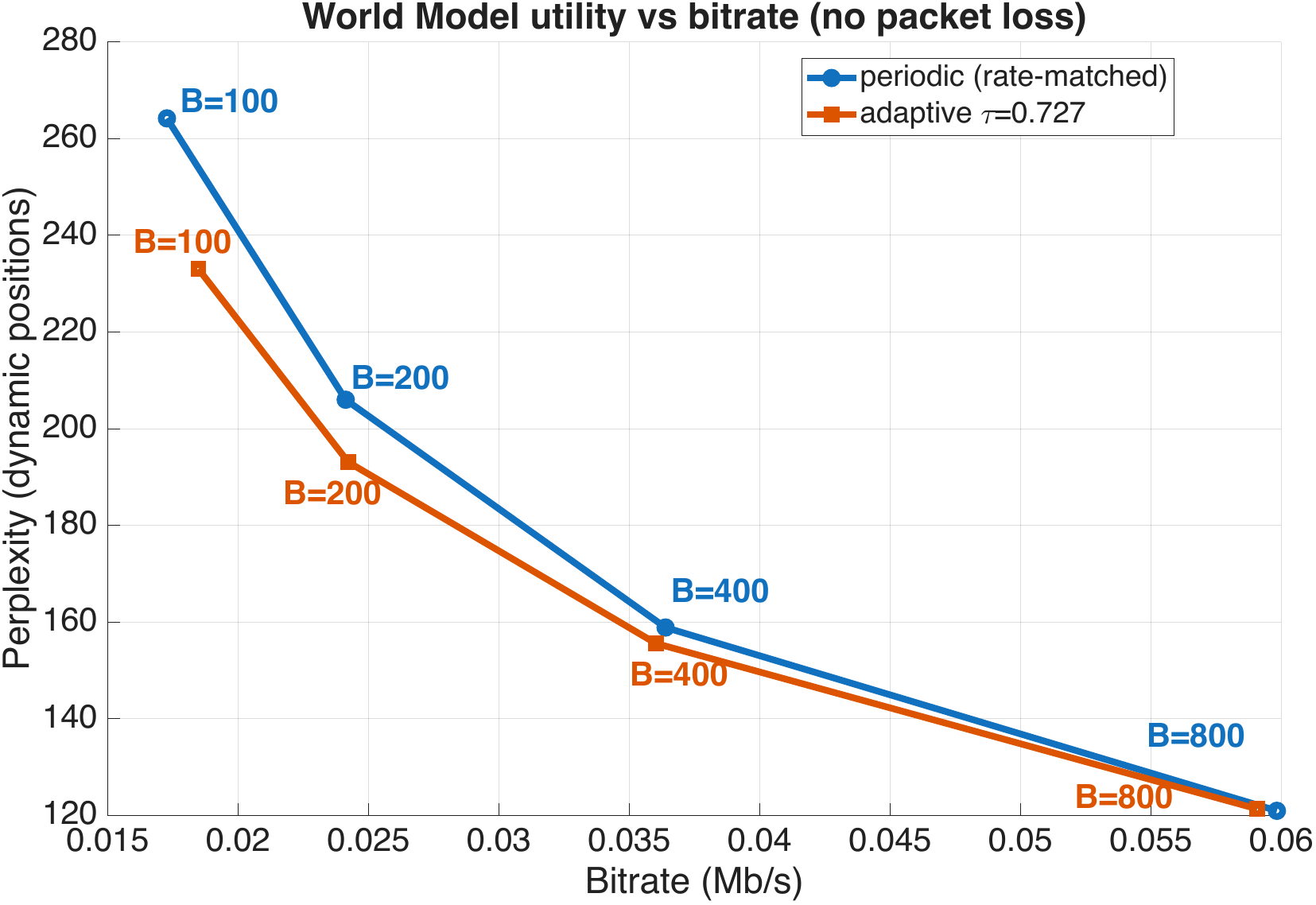}
  \caption{World model utility vs bitrate (no packet loss). Dynamic-position next-token perplexity (10\,Hz) for the adaptive algorithm ($\tau_h=0.727$, $N_{\max}=30$) versus rate-matched periodic baselines ($N=9$ near 200 bytes, $N=10$ near 400 bytes). Points are budgets $B\in\{100,200,400,800\}$ bytes (4 bytes/update); lower is better.}

  \label{fig:matlab_fig5_wm_utility_dyn_only}
\end{figure}

Finally, we connect token-state fidelity to downstream world model usefulness. We train a lightweight next-token predictor (context length $L{=}4$) on ground-truth token sequences, and evaluate its perplexity when conditioning on the receiver’s reconstructed token history $\hat{z}_{t-L:t-1}$. Figure \ref{fig:matlab_fig5_wm_utility_dyn_only} reports perplexity as a function of bitrate, for \emph{all positions} and \emph{dynamic positions} (positions that change between consecutive ground-truth frames). Perplexity decreases with bitrate for all methods, consistent with improved state synchronization yielding a more accurate conditioning context.

The most informative metric is dynamic-position perplexity, which isolates the part of the state most affected by missed updates. At the matched 200-byte operating point ($\sim$0.024\,Mb/s), adaptive keyframes reduce dynamic-position perplexity from 206.0 to 193.1 (6.3\% relative) compared to the matched periodic baseline. At the 400-byte operating point ($\sim$0.036\,Mb/s), adaptive keyframes reduce dynamic-position perplexity from 158.9 to 155.6 (2.1\% relative) against the matched periodic baseline. These improvements mirror the rate--distortion gains and indicate that the proposed streaming algorithm preserves token-dynamics utility where it matters most for next-step prediction.

\subsection{Summary}
Across fidelity, robustness, and utility evaluations, the proposed cosine-ranked delta selection and drift-triggered adaptive keyframes yield consistent improvements over periodic keyframes at matched bitrates, with the largest gains in the low-budget regime and on dynamic regions. Together the results support the conclusion that discrete token-state streaming can deliver bandwidth-aware synchronization while preserving downstream predictive usefulness under vehicular networking constraints.

\section{Discussion and Limitations}
Our results show that discrete token grids provide a practical interface for bandwidth-aware world-model state synchronization. The adaptive algorithm improves over periodic keyframes at matched bitrates, especially on dynamic regions where synchronization errors matter most, while gains narrow at higher bitrates as methods approach near-synchronization. We focus on dynamic-only embedding distortion to avoid static-region dominance and corroborate this metric with next-token perplexity conditioned on reconstructed receiver histories.

Our evaluation also has limitations relevant to deployment. The byte model uses a conservative fixed update cost (4 bytes/update plus a small header), abstracting away protocol engineering such as tighter bit packing or compression that would affect absolute bitrates. The network model assumes independent delta-message drops and reliable keyframes, whereas real vehicular links can exhibit bursty losses, delay and jitter, and may not guarantee keyframe delivery \cite{b7,b8}; evaluating on realistic channel traces and studying keyframe/delta protection is future work. Finally, the sender reference is maintained without feedback (assume delivered), and the utility probe is intentionally lightweight: it predicts each token using only its own temporal history at the same grid location (with a position embedding), without modeling spatial interactions across neighboring tokens. Stronger spatiotemporal predictors that exploit cross-position context and longer horizons may change absolute perplexity values and could amplify differences between streaming policies.

\section{Conclusion}
This work studied network-efficient synchronization of tokenized world model state for connected automated driving under payload and reliability constraints. Using a learned discrete tokenizer as the state representation, we proposed a keyframe--delta protocol with budgeted delta messages, where delta updates are selected online by cosine distance in codebook embedding space and keyframes are inserted adaptively based on a Hamming-drift trigger. Across $\sim$20\,s driving clips sampled at 10\,Hz, adaptive keyframes yield consistent rate-distortion improvements over periodic keyframes at matched bitrates: at $\sim$0.024\,Mb/s (200-byte budget) dynamic-only embedding distortion improves from 0.0712 to 0.0661, and at $\sim$0.036\,Mb/s (400-byte budget) from 0.0427 to 0.0407. Robustness experiments show distortion increases gradually with delta-message loss while keyframes bound desynchronization (e.g., at 10\% loss and 200 bytes, 0.0757 vs.\ 0.0789 for a matched periodic baseline), and a lightweight next-token predictor confirms preserved downstream utility (dynamic-position perplexity 206.0 to 193.1 at $\sim$0.024\,Mb/s). Overall, these results position discrete token-state streaming as a practical systems layer for scalable driving world models; future work will study tighter update compression, more realistic channel traces, and stronger spatiotemporal predictors for longer-horizon rollouts.

\end{document}